\documentclass[11pt,a4paper]{article}
\usepackage{emnlp2018}
\usepackage{times}
\usepackage{latexsym}

\usepackage{amsmath}
\usepackage{multirow}
\usepackage{url}

\usepackage{graphicx}
\usepackage{amsfonts}
\usepackage{bm}
\usepackage{microtype}

\aclfinalcopy 

\title{Direct Output Connection for a High-Rank Language Model}

\author{Sho Takase$^\dagger$ \hspace{1.5em} Jun Suzuki$^{\dagger \ddagger}$ \hspace{1.5em} Masaaki Nagata$^\dagger$ \\
  $\dagger$NTT Communication Science Laboratories \\ $\ddagger$Tohoku University \\
  {\tt \{takase.sho, nagata.masaaki\}@lab.ntt.co.jp}  \\ { \tt jun.suzuki@ecei.tohoku.ac.jp}
  }

\date{}

\begin{document}
\maketitle
\begin{abstract}
  This paper proposes a state-of-the-art recurrent neural network (RNN) language model that combines probability distributions computed not only from a final RNN layer but also from middle layers.
  Our proposed method raises the expressive power of a language model based on the matrix factorization interpretation of language modeling introduced by \protect\newcite{DBLP:journals/corr/abs-1711-03953}.
  The proposed method improves the current state-of-the-art language model and achieves the best score on the Penn Treebank and WikiText-2, which are the standard benchmark datasets.
  Moreover, we indicate our proposed method contributes to two application tasks: machine translation and headline generation.
  Our code is publicly available at: \href{https://github.com/nttcslab-nlp/doc_lm}{https://github.com/nttcslab-nlp/doc\_lm}.
\end{abstract}

\section{Introduction}
\label{sec:intro}
Neural network language models have played a central role in recent natural language processing (NLP) advances.
For example, neural encoder-decoder models, which were successfully applied to various natural language generation tasks including machine translation~\cite{Sutskever:2014:SSL:2969033.2969173}, summarization~\cite{rush-chopra-weston:2015:EMNLP}, and dialogue~\cite{wen-EtAl:2015:EMNLP}, can be interpreted as conditional neural language models.
Neural language models also positively influence syntactic parsing~\cite{dyer-EtAl:2016:N16-1,choe-charniak:2016:EMNLP2016}.
Moreover, such word embedding methods as Skip-gram~\cite{NIPS2013_5021} and vLBL~\cite{NIPS2013_5165} originated from neural language models designed to handle much larger vocabulary and data sizes.
Neural language models can also be used as contextualized word representations~\cite{N18-1202}.
Thus, language modeling is a good benchmark task for investigating the general frameworks of neural methods in NLP field.

In language modeling, we compute joint probability using the product of conditional probabilities.
Let $w_{1:T}$ be a word sequence with length $T$: $w_1, ..., w_T$.
We obtain the joint probability of word sequence $w_{1:T}$ as follows:
\begin{align}
  p(w_{1:T}) = p(w_1)\prod_{t=1}^{T-1} p(w_{t+1} | w_{1:t}). \label{eq:def_lm}
\end{align}
$p(w_1)$ is generally assumed to be $1$ in this literature, that is, $p(w_1)\!=\!1$, and thus we can ignore its calculation.
See the implementation of \newcite{DBLP:journals/corr/ZarembaSV14}\footnote{\href{https://github.com/wojzaremba/lstm}{{https://github.com/wojzaremba/lstm}}}, for an example.
RNN language models obtain conditional probability $p(w_{t+1} | w_{1:t})$ from the probability distribution of each word.
To compute the probability distribution, RNN language models encode sequence $w_{1:t}$ into a fixed-length vector and apply a transformation matrix and the softmax function.

Previous researches demonstrated that RNN language models achieve high performance by using several regularizations and selecting appropriate hyperparameters~\cite{DBLP:journals/corr/MelisDB17,merityRegOpt}.
However, \newcite{DBLP:journals/corr/abs-1711-03953} proved that existing RNN language models have low expressive power due to the \textit{Softmax bottleneck}, which means the output matrix of RNN language models is low rank when we interpret the training of RNN language models as a matrix factorization problem.
To solve the \textit{Softmax bottleneck}, \newcite{DBLP:journals/corr/abs-1711-03953} proposed \textit{Mixture of Softmaxes} (MoS), which increases the rank of the matrix by combining multiple probability distributions computed from the encoded fixed-length vector.

In this study, we propose \textit{Direct Output Connection} (DOC) as a generalization of MoS.
For stacked RNNs, DOC computes the probability distributions from the middle layers including input embeddings.
In addition to raising the rank, the proposed method helps weaken the vanishing gradient problem in backpropagation because DOC provides a shortcut connection to the output.

We conduct experiments on standard benchmark datasets for language modeling: the Penn Treebank and WikiText-2.
Our experiments demonstrate that DOC outperforms MoS and achieves state-of-the-art perplexities on each dataset.
Moreover, we investigate the effect of DOC on two applications: machine translation and headline generation.
We indicate that DOC can improve the performance of an encoder-decoder with an attention mechanism, which is a strong baseline for such applications.
In addition, we conduct an experiment on the Penn Treebank constituency parsing task to investigate the effectiveness of DOC.

\section{RNN Language Model}
In this section, we briefly overview RNN language models.
Let $V$ be the vocabulary size and let $P_{t} \in \mathbb{R}^{V}$ be the probability distribution of the vocabulary at timestep $t$.
Moreover, let $D_{h^n}$ be the dimension of the hidden state of the $n$-th RNN, and let $D_e$ be the dimensions of the embedding vectors.
Then the RNN language models predict probability distribution $P_{t+1}$ by the following equation:
\begin{align}
  P_{t+1} &= {\rm softmax}(W h^{N}_{t}), \label{eq:softmax} \\
  h^{n}_t &= f(h^{n-1}_{t}, h^{n}_{t-1}), \label{eq:rnn} \\
  h^{0}_t &= E x_t, \label{eq:embed}
\end{align}
where $W \in \mathbb{R}^{V \times D_{h^N}}$ is a weight matrix\footnote{Actually, we apply a bias term in addition to the weight matrix but we omit it to simplify the following discussion.}, $E \in \mathbb{R}^{D_e \times V}$ is a word embedding matrix, $x_t \in \{0,1\}^{V}$ is a one-hot vector of input word $w_t$ at timestep $t$, and $h^{n}_{t} \in \mathbb{R}^{D_{h^n}}$ is the hidden state of the $n$-th RNN at timestep $t$.
We define $h^{n}_{t}$ at timestep $t=0$ as a zero vector: $h^{n}_0 = \bm{0}$.
Let $f(\cdot)$ represent an abstract function of an RNN, which might be the Elman network~\cite{elman1990finding}, the Long Short-Term Memory (LSTM)~\cite{Hochreiter:1997:LSM:1246443.1246450}, the Recurrent Highway Network (RHN)~\cite{zilly2016recurrent}, or any other RNN variant.
In this research, we stack three LSTM layers based on \newcite{merityRegOpt} because they achieved high performance.

\section{Language Modeling as Matrix Factorization}
\label{sec:mos}
\newcite{DBLP:journals/corr/abs-1711-03953} indicated that the training of language models can be interpreted as a matrix factorization problem.
In this section, we briefly introduce their description.
Let word sequence $w_{1:t}$ be context $c_t$.
Then we can regard a natural language as a finite set of the pairs of a context and its conditional probability distribution: $\mathcal{L} = \{ (c_1, P^{*}(X|c_1)), ..., (c_U, P^{*}(X|c_U) ) \}$, where $U$ is the number of possible contexts and $X \in \{0, 1\}^{V}$ is a variable representing a one-hot vector of a word.
Here, we consider matrix $A \in \mathbb{R}^{U \times V}$ that represents the true log probability distributions and matrix $H \in \mathbb{R}^{U \times D_{h^N}}$ that contains the hidden states of the final RNN layer for each context $c_t$:
\begin{align}
  \label{eq:true_logp}
  A
  = 
  \begin{bmatrix}
  {\rm log} P^{*}(X | c_1) \\
  {\rm log} P^{*}(X | c_2) \\
  ... \\
  {\rm log} P^{*}(X | c_U) \\
  \end{bmatrix};
  H = 
  \begin{bmatrix}
  h^{N}_{c_1} \\
  h^{N}_{c_2} \\
  ... \\
  h^{N}_{c_U} \\
  \end{bmatrix}.
\end{align}
Then we obtain set of matrices $F(A) = \{A + \Lambda S \}$, where $S \in \mathbb{R}^{U \times V}$ is an all-ones matrix, and $\Lambda \in \mathbb{R}^{U \times U}$ is a diagonal matrix.
$F(A)$ contains matrices that shifted each row of $A$ by an arbitrary real number.
In other words, if we take a matrix from $F(A)$ and apply the softmax function to each of its rows, we obtain a matrix that consists of true probability distributions.
Therefore, for some $A' \in F(A)$, training RNN language models is to find the parameters satisfying the following equation:
\begin{align}
  H W^{\top} = A'. \label{eq:factorization}
\end{align}

Equation \ref{eq:factorization} indicates that training RNN language models can also be interpreted as a matrix factorization problem.
In most cases, the rank of matrix $H W^{\top}$ is $D_{h^N}$ because $D_{h^N}$ is smaller than $V$ and $U$ in common RNN language models.
Thus, an RNN language model cannot express true distributions if $D_{h^N}$ is much smaller than ${\rm rank}(A')$.

\newcite{DBLP:journals/corr/abs-1711-03953} also argued that ${\rm rank}(A')$ is as high as vocabulary size $V$ based on the following two assumptions:
\begin{enumerate}
  \item Natural language is highly context-dependent. In addition, since we can imagine many kinds of contexts, it is difficult to assume a basis that represents a conditional probability distribution for any contexts. In other words, compressing $U$ is difficult.
  \item Since we also have many kinds of semantic meanings, it is difficult to assume basic meanings that can create all other semantic meanings by such simple operations as addition and subtraction; compressing $V$ is difficult.
\end{enumerate}
In summary, \newcite{DBLP:journals/corr/abs-1711-03953} indicated that $D_{h^N}$ is much smaller than ${\rm rank}(A)$ because its scale is usually $10^2$ and vocabulary size $V$ is at least $10^4$.

\section{Proposed Method: Direct Output Connection}
\begin{figure}[!t]
  \centering
  \includegraphics[width=7cm]{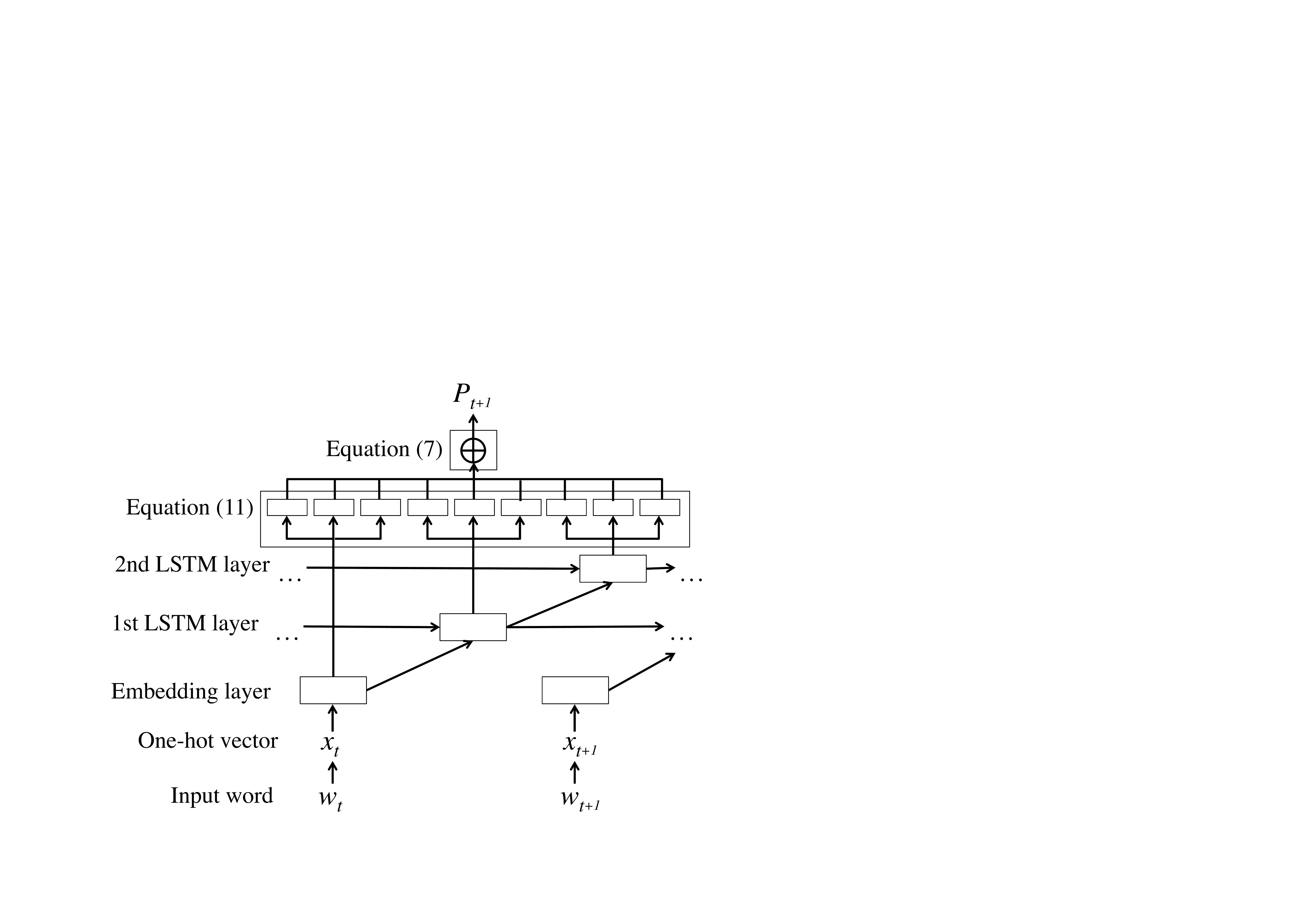}
   \caption{Overview of the proposed method: DOC. This figure represents the example of $N=2$ and $i_{0} = i_{1} = i_{2} = 3$.}
   \label{fig:overview}
\end{figure}

To construct a high-rank matrix, \newcite{DBLP:journals/corr/abs-1711-03953} proposed Mixture of Softmaxes (MoS).
MoS computes multiple probability distributions from the hidden state of final RNN layer $h^N$ and regards the weighted average of the probability distributions as the final distribution.
In this study, we propose Direct Output Connection (DOC), which is a generalization method of MoS.
DOC computes probability distributions from the middle layers in addition to the final layer.
In other words, DOC directly connects the middle layers to the output.

Figure \ref{fig:overview} shows an overview of DOC, that uses the middle layers (including word embeddings) to compute the probability distributions.
Figure \ref{fig:overview} computes three probability distributions from all the layers, but we can vary the number of probability distributions for each layer and select some layers to avoid.
In our experiments, we search for the appropriate number of probability distributions for each layer.

Formally, instead of Equation \ref{eq:softmax}, DOC computes the output probability distribution at timestep $t+1$ by the following equation:
\begin{align}
  P_{t+1} = &\sum_{j=1}^{J} \pi_{j, c_t} \  {\rm softmax} (\tilde{W} k_{j, c_t}), \label{eq:doc} \\
  s.t. &\sum_{j=1}^{J} \pi_{j, c_t} = 1,
\end{align}
where $\pi_{j, c_t}$ is a weight for each probability distribution, $k_{j, c_t} \in \mathbb{R}^{d}$ is a vector computed from each hidden state $h^{n}$, and $\tilde{W} \in \mathbb{R}^{V \times d}$ is a weight matrix.
Thus, $P_{t+1}$ is the weighted average of $J$ probability distributions.
We define the $U \times U$ diagonal matrix whose elements are weight $\pi_{j, c}$ for each context $c$ as $\Phi$.
Then we obtain matrix $\tilde{A} \in \mathbb{R}^{U \times V}$:
\begin{align}
  \tilde{A} = {\rm log} \sum_{j=1}^{J} \Phi \  {\rm softmax}(K_{j} \tilde{W}^\top), \label{eq:doc_factorization}
\end{align}
where $K_{j} \in \mathbb{R}^{U \times d}$ is a matrix whose rows are vector $k_{j, c}$.
$\tilde{A}$ can be an arbitrary high rank because the righthand side of Equation \ref{eq:doc_factorization} computes not only the matrix multiplication but also a nonlinear function.
Therefore, an RNN language model with DOC can output a distribution matrix whose rank is identical to one of the true distributions.
In other words, $\tilde{A}$ is a better approximation of $A'$ than the output of a standard RNN language model.

Next we describe how to acquire weight $\pi_{j, c_t}$ and vector $k_{j, c_t}$.
Let $\pi_{c_t} \in \mathbb{R}^{J}$ be a vector whose elements are weight $\pi_{j, c_t}$.
Then we compute $\pi_{c_t}$ from the hidden state of the final RNN layer:
\begin{align}
  \pi_{c_t} = {\rm softmax}(W_{\pi} h^{N}_{t} ), \label{eq:prior}
\end{align}
where $W_{\pi} \in \mathbb{R}^{J \times D_{h^N}}$ is a weight matrix.
We next compute $k_{j, c_t}$ from the hidden state of the $n$-th RNN layer:
\begin{align}
  k_{j, c_t} = W_j h^{n}_t,
\end{align}
where $W_j \in \mathbb{R}^{d \times D_{h^n}}$ is a weight matrix.
In addition, let $i_n$ be the number of $k_{j, c_t}$ from $h^{n}_t$.
Then we define the sum of $i_n$ for all $n$ as $J$; that is, $\sum_{n=0}^N i_n = J$.
In short, DOC computes $J$ probability distributions from all the layers, including the input embedding ($h^0$).
For $i_N = J$, DOC becomes identical to MoS.
In addition to increasing the rank, we expect that DOC weakens the vanishing gradient problem during backpropagation because a middle layer is directly connected to the output, such as with the auxiliary classifiers described in \newcite{43022}.

For a network that computes the weights for several vectors, such as Equation \ref{eq:prior}, \newcite{DBLP:journals/corr/ShazeerMMDLHD17} indicated that it often converges to a state where it always produces large weights for few vectors.
In fact, we observed that DOC tends to assign large weights to shallow layers.
To prevent this phenomenon, we compute the coefficient of variation of Equation \ref{eq:prior} in each mini-batch as a regularization term following \newcite{DBLP:journals/corr/ShazeerMMDLHD17}.
In other words, we try to adjust the sum of the weights for each probability distribution with identical values in each mini-batch.
Formally, we compute the following equation for a mini-batch consisting of $w_{b}, w_{b+1}, ..., w_{\tilde{b}}$:
\begin{align}
  B &= \sum_{t=b}^{\tilde{b}} \pi_{c_{t}} \\
  \beta &= \biggl(\frac{{\rm std}(B)} {{\rm avg}(B)} \biggr)^2, \label{eq:cv}
\end{align}
where functions ${\rm std}(\cdot)$ and ${\rm avg}(\cdot)$ are functions that respectively return an input's standard deviation and its average.
In the training step, we add $\lambda_{\beta}$ multiplied by weight coefficient $\beta$ to the loss function.

\section{Experiments on Language Modeling}
\label{sec:lm_exp}
We investigate the effect of DOC on the language modeling task.
In detail, we conduct word-level prediction experiments and show that DOC improves the performance of MoS, which only uses the final layer to compute the probability distributions.
Moreover, we evaluate various combinations of layers to explore which combination achieves the best score.

\subsection{Datasets}
We used the Penn Treebank (PTB)~\cite{Marcus:1993:BLA:972470.972475} and WikiText-2~\cite{DBLP:journals/corr/MerityXBS16} datasets, which are the standard benchmark datasets for the word-level language modeling task.
\newcite{DBLP:conf/interspeech/MikolovKBCK10} and \newcite{DBLP:journals/corr/MerityXBS16} respectively published preprocessed PTB\footnote{\href{http://www.fit.vutbr.cz/~imikolov/rnnlm/}{{http://www.fit.vutbr.cz/~imikolov/rnnlm/}}} and WikiText-2\footnote{\href{https://einstein.ai/research/the-wikitext-long-term-dependency-language-modeling-dataset}{{https://einstein.ai/research/the-wikitext-long-term-dependency-language-modeling-dataset}}} datasets.
Table \ref{tab:dataset} describes their statistics.
We used these preprocessed datasets for fair comparisons with previous studies.

\subsection{Hyperparameters}
Our implementation is based on the averaged stochastic gradient descent Weight-Dropped LSTM (AWD-LSTM)\footnote{\href{https://github.com/salesforce/awd-lstm-lm}{https://github.com/salesforce/awd-lstm-lm}} proposed by \newcite{merityRegOpt}.
AWD-LSTM consists of three LSTMs with various regularizations.
For the hyperparameters, we used the same values as \newcite{DBLP:journals/corr/abs-1711-03953} except for the dropout rate for vector $k_{j, c_t}$ and the non-monotone interval.
Since we found that the dropout rate for vector $k_{j, c_t}$ greatly influences $\beta$ in Equation \ref{eq:cv}, we varied it from $0.3$ to $0.6$ with $0.1$ intervals.
We selected $0.6$ because this value achieved the best score on the PTB validation dataset.
For the non-monotone interval, we adopted the same value as \newcite{fraternal}.
Table \ref{tab:hyperparameters} summarizes the hyperparameters of our experiments.

\begin{table}[!t]
  \centering
  \small
  \begin{tabular}{| c c | r | r |} \hline
  \multicolumn{2}{|c|}{} & \multicolumn{1}{c|}{PTB} & \multicolumn{1}{c|}{WikiText-2} \\ \hline
  \multicolumn{2}{|c|}{Vocab} & 10,000 & 33,278 \\ \hline
   & Train & 929,590 & 2,088,628 \\
   \#Token & Valid & 73,761 & 217,646 \\
   & Test & 82,431 & 245,569 \\ \hline
  \end{tabular}
  \caption{Statistics of PTB and WikiText-2.\label{tab:dataset}}
\end{table}

\begin{table}[!t]
  \centering
  \small
  \begin{tabular}{| l | r | r |} \hline
  Hyperparameter & PTB & WikiText-2 \\ \hline
  Learning rate & 20 & 15 \\
  Batch size & 12 & 15 \\
  Non-monotone interval & 60 & 60 \\
  $D_e$ & 280 & 300 \\
  $D_{h^1}$ & 960 & 1150 \\
  $D_{h^2}$ & 960 & 1150 \\
  $D_{h^3}$ & 620 & 650 \\ \hline
  Dropout rate for $x_t$ & 0.1 & 0.1 \\
  Dropout rate for $h^0_t$ & 0.4 & 0.65 \\
  Dropout rate for $h^1_t, h^2_t$ & 0.225 & 0.2 \\
  Dropout rate for $h^3_t$ & 0.4 & 0.4 \\
  Dropout rate for $k_{j, c_t}$ & 0.6 & 0.6 \\
  Recurrent weight dropout & 0.50 & 0.50 \\ \hline
  \end{tabular}
  \caption{Hyperparameters used for training DOC.\label{tab:hyperparameters}}
\end{table}

\begin{table}[!t]
  \centering
  \small
  \begin{tabular}{| c c c c | r | c c |} \hline
  \multicolumn{4}{|c|}{\#DOC} & & & \\
  $i_3$ & $i_2$ & $i_1$ & $i_0$ & \multicolumn{1}{c|}{ $\lambda_{\beta}$ } & Valid & Test \\ \hline
  15 & 0 & 0 & 0 & 0 & 56.54$\dagger$ & 54.44$\dagger$ \\ \hline
  20 & 0 & 0 & 0 & 0 & 56.88$\ddagger$ & 54.79$\ddagger$ \\
  15 & 0 & 0 & 5 & 0 & 56.21 & 54.28 \\
  15 & 0 & 5 & 0 & 0 & 55.26 & 53.52 \\
  15 & 5 & 0 & 0 & 0 & 54.87 & 53.15 \\
  15 & 5 & 0 & 0 & 0.0001 & 54.95 & 53.16 \\
  15 & 5 & 0 & 0 & 0.001 & {\bf 54.62} & {\bf 52.87} \\
  15 & 5 & 0 & 0 & 0.01 & 55.13 & 53.39 \\
  10 & 5 & 0 & 5 & 0 & 56.46 & 54.18 \\
  10 & 5 & 5 & 0 & 0 & 56.00 & 54.37 \\ \hline
  \end{tabular}
  \caption{Perplexities of AWD-LSTM with DOC on the PTB dataset. We varied the number of probability distributions from each layer in situation $J=20$ except for the top row. The top row ($\dagger$) represents MoS scores reported in \protect\newcite{DBLP:journals/corr/abs-1711-03953} as a baseline. $\ddagger$ represents the perplexity obtained by the implementation of \protect\newcite{DBLP:journals/corr/abs-1711-03953}\protect\footnotemark[6] with identical hyperparameters except for $i_3$. \label{tb:search_doc}}
\end{table}
\addtocounter{footnote}{1}
\footnotetext{\href{https://github.com/zihangdai/mos}{https://github.com/zihangdai/mos}}

\subsection{Results}
\label{Sec:lang_result}

Table \ref{tb:search_doc} shows the perplexities of AWD-LSTM with DOC on the PTB dataset.
Each value of columns $i_n$ represents the number of probability distributions from hidden state $h^n_t$.
To find the best combination, we varied the number of probability distributions from each layer by fixing their total to 20: $J=20$.
Moreover, the top row of Table \ref{tb:search_doc} shows the perplexity of AWD-LSTM with MoS reported in \newcite{DBLP:journals/corr/abs-1711-03953} for comparison.
Table \ref{tb:search_doc} indicates that language models using middle layers outperformed one using only the final layer.
In addition, Table \ref{tb:search_doc} shows that increasing the distributions from the final layer ($i_{3}=20$) degraded the score from the language model with $i_{3}=15$ (the top row of Table \ref{tb:search_doc}).
Thus, to obtain a superior language model, we should not increase the number of distributions from the final layer; we should instead use the middle layers, as with our proposed DOC.

\begin{table}[!t]
  \centering
  \small
  \begin{tabular}{| r | c c |} \hline
  \multicolumn{1}{|c|}{ $\lambda_{\beta}$ } & Valid & Test \\ \hline
  0 & 0.276 & 0.279 \\
  0.0001 & 0.254 & 0.252 \\
  0.001 & 0.217 & 0.213 \\
  0.01 & 0.092 & 0.086 \\ \hline
  \end{tabular}
  \caption{Coefficient of variation of Equation \ref{eq:prior}: $\sqrt{\beta}$ in validation and test sets of PTB.\label{tb:cv}}
\end{table}

\begin{table}[!t]
  \centering
  \small
  \begin{tabular}{| l | r r |} \hline
  Model & Valid & Test \\ \hline
  AWD-LSTM & 401 & 401 \\
  AWD-LSTM-MoS & 10000 & 10000 \\
  AWD-LSTM-DOC & 10000 & 10000 \\ \hline
  \end{tabular}
  \caption{Rank of output matrix ($\tilde{A}$ in Equation \ref{eq:doc_factorization}) on the PTB dataset. $D_{3}$ of AWD-LSTM is 400.\label{tb:output_rank}}
\end{table}

Table \ref{tb:search_doc} shows that the $i_{3}=15, i_{2}=5$ setting achieved the best performance and the other settings with shallow layers have a little effect.
This result implies that we need some layers to output accurate distributions.
In fact, most previous studies adopted two LSTM layers for language modeling.
This suggests that we need at least two layers to obtain high-quality distributions.

\begin{figure}[!t]
  \centering
  \includegraphics[width=7cm]{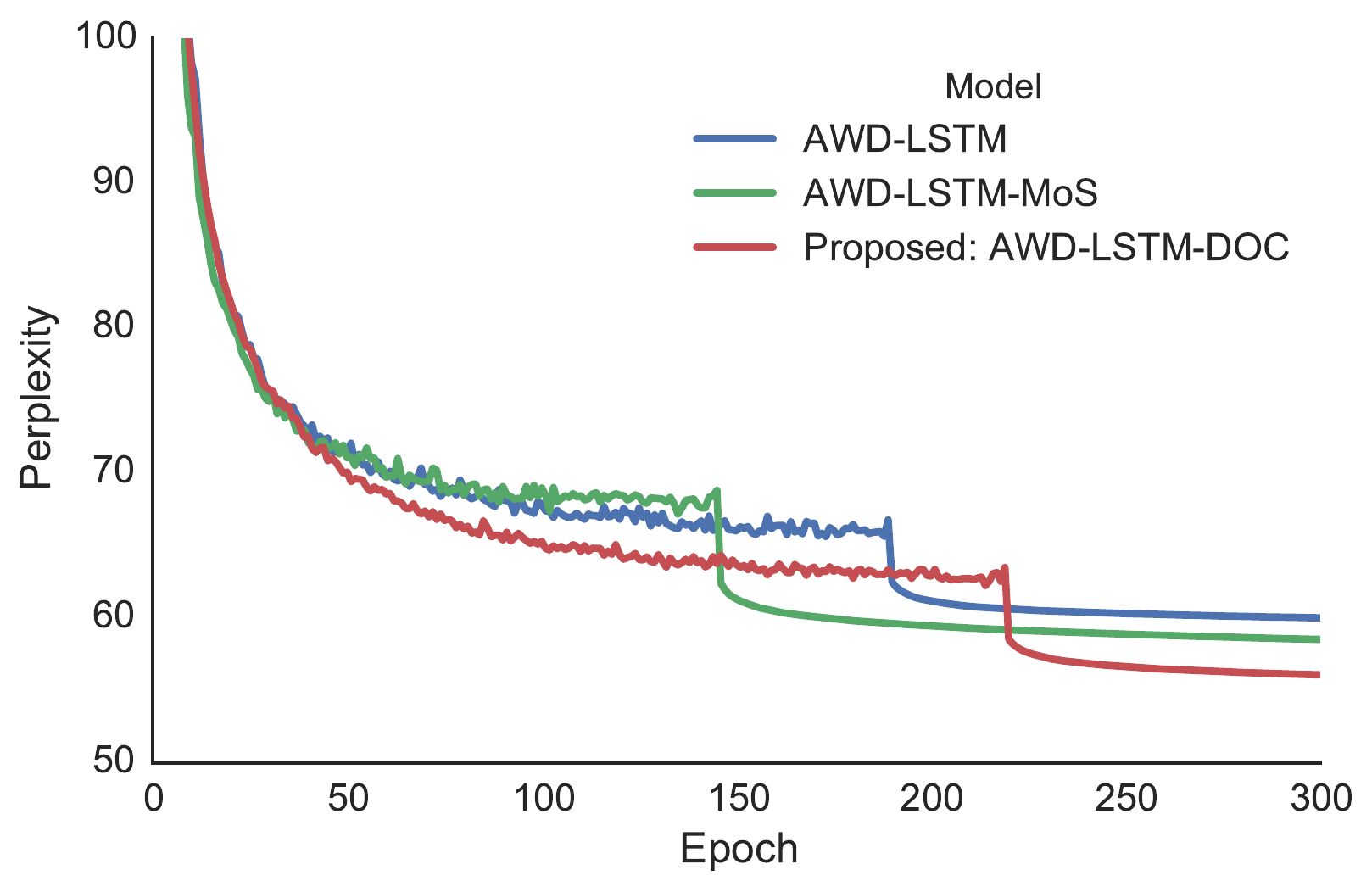}
   \caption{Perplexities of each method on the PTB validation set.\label{fig:learning_curve}}
\end{figure}

\begin{table}[!t]
  \centering
  \small
  \begin{tabular}{| l | r r |} \hline
  Model & Valid & Test \\ \hline
  AWD-LSTM$\dagger$ & 58.88 & 56.36 \\
  AWD-LSTM-MoS$\dagger$ & 56.36 & 54.26 \\
  AWD-LSTM-MoS$\ddagger$ & 55.67 & 53.75 \\
  AWD-LSTM-DOC & {\bf 54.62} & {\bf 52.87} \\
  AWD-LSTM-DOC (fin) & {\bf 54.12} & {\bf 52.38} \\ \hline
  \end{tabular}
  \caption{Perplexities of our implementations and re-runs on the PTB dataset. We set the non-monotone interval to 60. $\dagger$ represents results obtained by original implementations with identical hyperparameters except for non-monotone interval. $\ddagger$ indicates the result obtained by our AWD-LSTM-MoS implementation with identical dropout rates as AWD-LSTM-DOC. For (fin), we repeated fine-tuning until convergence.\label{tb:compare_awdlstm}}
\end{table}

For the $i_{3}=15, i_{2}=5$ setting, we explored the effect of $\lambda_{\beta}$ in $\{0, 0.01, 0.001, 0.0001\}$.
Although Table \ref{tb:search_doc} shows that $\lambda_{\beta} = 0.001$ achieved the best perplexity, the effect is not consistent.
Table \ref{tb:cv} shows the coefficient of variation of Equation \ref{eq:prior}, i.e., $\sqrt{\beta}$ in the PTB dataset.
This table demonstrates that the coefficient of variation decreases with growth in $\lambda_{\beta}$.
In other words, the model trained with a large $\lambda_{\beta}$ assigns balanced weights to each probability distribution.
These results indicate that it is not always necessary to equally use each probability distribution, but we can acquire a better model in some $\lambda_{\beta}$.
Hereafter, we refer to the setting that achieved the best score ($i_{3}=15, i_{2}=5, \lambda_{\beta}=0.001$) as AWD-LSTM-DOC.

\begin{table*}[!t]
  \centering
  \small
  \begin{tabular}{| l  | c | c c |} \hline
  Model & \#Param & Valid & Test \\ \hline
  LSTM (medium) \cite{DBLP:journals/corr/ZarembaSV14} & 20M & 86.2  \ \  & 82.7  \ \  \\
  LSTM (large) \cite{DBLP:journals/corr/ZarembaSV14} & 66M & 82.2  \ \  & 78.4  \ \  \\
  Variational LSTM (medium) \cite{Gal2016Theoretically} & 20M & 81.9 $\pm$ 0.2 & 79.7 $\pm$ 0.1 \\
  Variational LSTM (large) \cite{Gal2016Theoretically} & 66M & 77.9 $\pm$ 0.3 & 75.2 $\pm$ 0.2 \\
  Variational RHN \cite{zilly2016recurrent} & 32M & 71.2  \ \  & 68.5  \ \  \\
  Variational RHN + WT \cite{zilly2016recurrent} & 23M & 67.9  \ \  & 65.4  \ \ \\
  Variational RHN + WT + IOG \cite{takase-suzuki-nagata:2017:I17-2} & 29M & 67.0  \ \  & 64.4 \ \  \\
  Neural Architecture Search \cite{45826} & 54M & - & 62.4 \ \  \\
  LSTM with skip connections \cite{DBLP:journals/corr/MelisDB17} & 24M & 60.9  \ \  & 58.3 \ \  \\
  AWD-LSTM \cite{merityRegOpt} & 24M & 60.0  \ \  & 57.3 \ \  \\
  AWD-LSTM + Fraternal Dropout \cite{fraternal} & 24M & 58.9  \ \  & 56.8 \ \  \\
  AWD-LSTM-MoS \cite{DBLP:journals/corr/abs-1711-03953} & 22M & 56.54 & 54.44 \\ \hline
  Proposed method: AWD-LSTM-DOC & 23M & {\bf 54.62} & {\bf 52.87} \\
  Proposed method: AWD-LSTM-DOC (fin) & 23M & {\bf 54.12} & {\bf 52.38} \\
  Proposed method (ensemble): AWD-LSTM-DOC $\times$ 5 & 114M & {\bf 49.99} & {\bf 48.44} \\
  Proposed method (ensemble): AWD-LSTM-DOC (fin) $\times$ 5 & 114M & {\bf 48.63} & {\bf 47.17} \\ \hline
  \end{tabular}
  \caption{Perplexities of each method on the PTB dataset.\label{tb:perplexity}}
\end{table*}

\begin{table*}[!t]
  \centering
  \small
  \begin{tabular}{| l | c | c c |} \hline
  Model & \#Param & Valid & Test \\ \hline
  Variational LSTM + IOG \cite{takase-suzuki-nagata:2017:I17-2} & 70M & 95.9  \ \  & 91.0 \ \  \\
  Variational LSTM + WT + AL \cite{DBLP:journals/corr/InanKS16} & 28M & 91.5  \ \ & 87.0 \ \  \\
  LSTM with skip connections \cite{DBLP:journals/corr/MelisDB17} & 24M & 69.1  \ \  & 65.9 \ \  \\
  AWD-LSTM \cite{merityRegOpt} & 33M & 68.6  \ \  & 65.8 \ \  \\
  AWD-LSTM + Fraternal Dropout \cite{fraternal} & 34M & 66.8  \ \  & 64.1 \ \  \\
  AWD-LSTM-MoS \cite{DBLP:journals/corr/abs-1711-03953} & 35M & 63.88 & 61.45 \\ \hline
  Proposed method: AWD-LSTM-DOC & 37M & {\bf 60.97} & {\bf 58.55} \\
  Proposed method: AWD-LSTM-DOC (fin) & 37M & {\bf 60.29} & {\bf 58.03} \\
  Proposed method (ensemble): AWD-LSTM-DOC $\times$ 5 & 185M & {\bf 56.14} & {\bf 54.23} \\
  Proposed method (ensemble): AWD-LSTM-DOC (fin) $\times$ 5 & 185M & {\bf 54.91} & {\bf 53.09} \\ \hline
  \end{tabular}
  \caption{Perplexities of each method on the WikiText-2 dataset.\label{tb:perplexityOnWikitext}}
\end{table*}

Table \ref{tb:output_rank} shows the ranks of matrices containing log probability distributions from each method.
In other words, Table \ref{tb:output_rank} describes $\tilde{A}$ in Equation \ref{eq:doc_factorization} for each method.
As shown by this table, the output of AWD-LSTM is restricted to $D_3$\footnote{Actually, the maximum rank size of an ordinary RNN language model is $D_{N} + 1$ when we use a bias term.}.
In contrast, AWD-LSTM-MoS~\cite{DBLP:journals/corr/abs-1711-03953} and AWD-LSTM-DOC outputted matrices whose ranks equal the vocabulary size.
This fact indicates that DOC (including MoS) can output the same matrix as the true distributions in view of a rank.

Figure \ref{fig:learning_curve} illustrates the learning curves of each method on PTB.
This figure contains the validation scores of AWD-LSTM, AWD-LSTM-MoS, and AWD-LSTM-DOC at each training epoch.
We trained AWD-LSTM and AWD-LSTM-MoS by setting the non-monotone interval to 60, as with AWD-LSTM-DOC.
In other words, we used hyperparameters identical to the original ones to train AWD-LSTM and AWD-LSTM-MoS, except for the non-monotone interval.
We note that the optimization method converts the ordinary stochastic gradient descent (SGD) into the averaged SGD at the point where convergence almost occurs.
In Figure \ref{fig:learning_curve}, the turning point is the epoch when each method drastically decreases the perplexity.
Figure \ref{fig:learning_curve} shows that each method similarly reduces the perplexity at the beginning.
AWD-LSTM and AWD-LSTM-MoS were slow to decrease the perplexity from 50 epochs.
In contrast, AWD-LSTM-DOC constantly decreased the perplexity and achieved a lower value than the other methods with ordinary SGD.
Therefore, we conclude that DOC positively affects the training of language modeling.

Table \ref{tb:compare_awdlstm} shows the AWD-LSTM, AWD-LSTM-MoS, and AWD-LSTM-DOC results in our configurations.
For AWD-LSTM-MoS, we trained our implementation with the same dropout rates as AWD-LSTM-DOC for a fair comparison.
AWD-LSTM-DOC outperformed both the original AWD-LSTM-MoS and our implementation.
In other words, DOC outperformed MoS.

Since the averaged SGD uses the averaged parameters from each update step, the parameters of the early steps are harmful to the final parameters.
Therefore, when the model converges, recent studies and ours eliminate the history of and then retrains the model.
\newcite{merityRegOpt} referred to this retraining process as fine-tuning.
Although most previous studies only conducted fine-tuning once, \newcite{fraternal} argued that two fine-tunings provided additional improvement.
Thus, we repeated fine-tuning until we achieved no more improvements in the validation data.
We refer to the model as AWD-LSTM-DOC (fin) in Table \ref{tb:compare_awdlstm}, which shows that repeated fine-tunings improved the perplexity by about 0.5.

Tables \ref{tb:perplexity} and \ref{tb:perplexityOnWikitext} respectively show the perplexities of AWD-LSTM-DOC and previous studies on PTB and WikiText-2\footnote{We exclude models that use the statistics of the test data~\cite{DBLP:journals/corr/GraveJU16,DBLP:journals/corr/abs-1709-07432} from these tables because we regard neural language models as the basis of NLP applications and consider it unreasonable to know correct outputs during applications, e.g., machine translation. In other words, we focus on neural language models as the foundation of applications although we can combine the method using the statistics of test data with our AWD-LSTM-DOC.}.
These tables show that AWD-LSTM-DOC achieved the best perplexity.
AWD-LSTM-DOC improved the perplexity by almost 2.0 on PTB and 3.5 on WikiText-2 from the state-of-the-art scores.
The ensemble technique provided further improvement, as described in previous studies~\cite{DBLP:journals/corr/ZarembaSV14,takase-suzuki-nagata:2017:I17-2}, and improved the perplexity by at least 4 points on both datasets.
Finally, the ensemble of the repeated finetuning models achieved 47.17 on the PTB test and 53.09 on the WikiText-2 test.

\section{Experiments on Application Tasks}
\label{sec:exp_generation}

As described in Section \ref{sec:intro}, a neural encoder-decoder model can be interpreted as a conditional language model.
To investigate the effect of DOC on an encoder-decoder model, we incorporate DOC into the decoder and examine its performance.

\subsection{Dataset}
We conducted experiments on machine translation and headline generation tasks.
For machine translation, we used two kinds of sentence pairs (English-German and English-French) in the IWSLT 2016 dataset\footnote{\href{https://wit3.fbk.eu/}{https://wit3.fbk.eu/}}.
The training set respectively contains about 189K and 208K sentence pairs of English-German and English-French.
We experimented in four settings: from English to German (En-De), its reverse (De-En), from English to French (En-Fr), and its reverse (Fr-En).

Headline generation is a task that creates a short summarization of an input sentence\cite{rush-chopra-weston:2015:EMNLP}.
\newcite{rush-chopra-weston:2015:EMNLP} constructed a headline generation dataset by extracting pairs of first sentences of news articles and their headlines from the annotated English Gigaword corpus~\cite{Napoles:2012:AG:2391200.2391218}.
They also divided the extracted sentence-headline pairs into three parts: training, validation, and test sets.
The training set contains about 3.8M sentence-headline pairs.
For our evaluation, we used the test set constructed by \newcite{zhou-EtAl:2017:Long} because the one constructed by \newcite{rush-chopra-weston:2015:EMNLP} contains some invalid instances, as reported in \newcite{zhou-EtAl:2017:Long}.

\subsection{Encoder-Decoder Model}
For the base model, we adopted an encoder-decoder with an attention mechanism described in \newcite{kiyono}.
The encoder consists of a 2-layer bidirectional LSTM, and the decoder consists of a 2-layer LSTM with attention proposed by \newcite{luong-pham-manning:2015:EMNLP}.
We interpreted the layer after computing the attention as the 3rd layer of the decoder.
We refer to this encoder-decoder as EncDec.
For the hyperparameters, we followed the setting of \newcite{kiyono} except for the sizes of hidden states and embeddings.
We used 500 for machine translation and 400 for headline generation.
We constructed a vocabulary set by using Byte-Pair-Encoding\footnote{\href{https://github.com/rsennrich/subword-nmt}{https://github.com/rsennrich/subword-nmt}} (BPE)~\cite{sennrich-haddow-birch:2016:P16-11}.
We set the number of BPE merge operations at 16K for the machine translation and 5K for the headline generation.

In this experiment, we compare DOC to the base EncDec.
We prepared two DOC settings: using only the final layer, that is, a setting that is identical to MoS, and using both the final and middle layers.
We used the 2nd and 3rd layers in the latter setting because this case achieved the best performance on the language modeling task in Section \ref{Sec:lang_result}.
We set $i_{3} = 2$ and $i_{2} = 2, i_{3} = 2$.
For this experiment, we modified a publicly available encode-decoder implementation\footnote{\href{https://github.com/mlpnlp/mlpnlp-nmt/}{https://github.com/mlpnlp/mlpnlp-nmt/}}.

\subsection{Results}

\begin{table}[!t]
  \centering
  \small
  \tabcolsep=2.5pt
  \begin{tabular}{| l | c c c c |} \hline
  Model & En-De & De-En & En-Fr & Fr-En\\ \hline
  EncDec & 23.05 & 28.18 & 34.37 & 34.07 \\
  EncDec+DOC ($i_3 = 2$) & 23.62 & 29.12 & 36.09 & 34.41 \\
  EncDec+DOC ($i_3 = i_2 = 2$) & {\bf 23.97} & {\bf 29.33} & {\bf 36.11} & {\bf 34.72} \\ \hline
  \end{tabular}
  \caption{BLEU scores on test sets in the IWSLT 2016 dataset. We report averages of three runs.\label{tb:nmt}}
\end{table}

\begin{table}[!t]
  \centering
  \small
  \begin{tabular}{| l | c c c |} \hline
  Model & RG-1 & RG-2 & RG-L \\ \hline
  EncDec & 46.77 & 24.87 & 43.58 \\
  EncDec+DOC ($i_3 = 2$) & 46.91 & 24.91 & 43.73 \\
  EncDec+DOC ($i_3 = i_2 = 2$) & {\bf 46.99} & {\bf 25.29} & {\bf 43.83} \\ \hline
  ABS \cite{rush-chopra-weston:2015:EMNLP} & 37.41 & 15.87 & 34.70 \\
  SEASS \cite{zhou-EtAl:2017:Long} & 46.86 & 24.58 & 43.53 \\ 
  \newcite{kiyono} & 46.34 & 24.85 & 43.49  \\ \hline
  \end{tabular}
  \caption{ROUGE F1 scores in headline generation test data provided by \protect\newcite{zhou-EtAl:2017:Long}. RG in table denotes ROUGE. For our implementations (the upper part), we report averages of three runs.\label{tb:headline_zhou}}
\end{table}

Table \ref{tb:nmt} shows the BLEU scores of each method.
Since an initial value often drastically varies the result of a neural encoder-decoder, we reported the average of three models trained from different initial values and random seeds.
Table \ref{tb:nmt} indicates that EncDec+DOC outperformed EncDec.

Table \ref{tb:headline_zhou} shows the ROUGE F1 scores of each method.
In addition to the results of our implementations (the upper part), the lower part represents the published scores reported in previous studies.
For the upper part, we reported the average of three models (as in Table \ref{tb:nmt}).
EncDec+DOC outperformed EncDec on all scores.
Moreover, EncDec outperformed the state-of-the-art method~\cite{zhou-EtAl:2017:Long} on the ROUGE-2 and ROUGE-L F1 scores.
In other words, our baseline is already very strong.
We believe that this is because we adopted a larger embedding size than \newcite{zhou-EtAl:2017:Long}.
It is noteworthy that DOC improved the performance of EncDec even though EncDec is very strong.

These results indicate that DOC positively influences a neural encoder-decoder model.
Using the middle layer also yields further improvement because EncDec+DOC ($i_3 = i_2 = 2$) outperformed EncDec+DOC ($i_3 = 2$).

\section{Experiments on Constituency Parsing} \label{sec:exp_in_parsing}
\newcite{choe-charniak:2016:EMNLP2016} achieved high F1 scores on the Penn Treebank constituency parsing task by transforming candidate trees into a symbol sequence (S-expression) and reranking them based on the perplexity obtained by a neural language model.
To investigate the effectiveness of DOC, we evaluate our language models following their configurations.

\subsection{Dataset}
We used the Wall Street Journal of the Penn Treebank dataset.
We used the section 2-21 for training, 22 for validation, and 23 for testing.
We applied the preprocessing codes of \newcite{choe-charniak:2016:EMNLP2016}\footnote{\href{https://github.com/cdg720/emnlp2016}{https://github.com/cdg720/emnlp2016}} to the dataset and converted a token that appears fewer than ten times in the training dataset into a special token {\it unk}.
For reranking, we prepared 500 candidates obtained by the Charniak parser~\cite{A00-2018}.

\subsection{Models}
We compare AWD-LSTM-DOC with AWD-LSTM~\cite{merityRegOpt} and AWD-LSTM-MoS~\cite{DBLP:journals/corr/abs-1711-03953}.
We trained each model with the same hyperparameters from our language modeling experiments (Section \ref{sec:lm_exp}).
We selected the model that achieved the best perplexity on the validation set during the training.

\subsection{Results}
\begin{table}[!t]
  \centering
  \small
  \tabcolsep=5pt
  \begin{tabular}{| l | c c |} \hline
  & \multicolumn{2}{c|}{F1} \\
  Model & Base & Rerank \\ \hline
  \multicolumn{3}{|c|}{Reranking with single model} \\ \hline
  \newcite{choe-charniak:2016:EMNLP2016} & 89.7 & 92.6 \\
  AWD-LSTM & 89.7  & 93.2 \\
  AWD-LSTM-MoS & 89.7  & 93.2 \\
  AWD-LSTM-DOC & 89.7 & {\bf 93.3} \\ \hline
  \multicolumn{3}{|c|}{Reranking with model ensemble} \\ \hline
  AWD-LSTM $\times$ 5 (ensemble) & 89.7 \ \  & 93.4  \ \  \\
  AWD-LSTM-MoS $\times$ 5 (ensemble) & 89.7 \ \  & 93.4 \ \  \\
  AWD-LSTM-DOC $\times$ 5 (ensemble) & 89.7 \ \  & {\bf 93.5}  \ \  \\
  AWD-LSTM-DOC $\times$ 5 (ensemble) & 91.2 \ \  & {\bf 94.29} \\
  AWD-LSTM-DOC $\times$ 5 (ensemble) & 93.12 & {\bf 94.47} \\ \hline
  \multicolumn{3}{|c|}{State-of-the-art results} \\ \hline
  \newcite{dyer-EtAl:2016:N16-1} & 91.7 \ \ & 93.3  \ \  \\
  \newcite{fried-stern-klein:2017:Short} (ensemble) & 92.72 & 94.25 \\
  \newcite{P18-2097} (ensemble) & 92.74 & 94.32 \\
  \newcite{P18-1249} & 95.13 & - \\ \hline
  \end{tabular}
  \caption{Bracketing F1 scores on the PTB test set (Section 23). This table includes reranking models trained on the PTB without external data.\label{tb:parse_result}}
\end{table}

Table \ref{tb:parse_result} shows the bracketing F1 scores on the PTB test set.
This table is divided into three parts by horizontal lines; the upper part describes the scores by single language modeling based rerankers, the middle part shows the results by ensembling five rerankers, and the lower part represents the current state-of-the-art scores in the setting without external data.
The upper part also contains the score reported in \newcite{choe-charniak:2016:EMNLP2016} that reranked candidates by the simple LSTM language model.
This part indicates that our implemented rerankers outperformed the simple LSTM language model based reranker, which achieved 92.6 F1 score~\cite{choe-charniak:2016:EMNLP2016}.
Moreover, AWD-LSTM-DOC outperformed AWD-LSTM and AWD-LSTM-MoS.
These results correspond to the performance on the language modeling task (Section \ref{Sec:lang_result}).

The middle part shows that AWD-LSTM-DOC also outperformed AWD-LSTM and AWD-LSTM-MoS in the ensemble setting.
In addition, we can improve the performance by exchanging the base parser with a stronger one.
In fact, we achieved 94.29 F1 score by reranking the candidates from retrained Recurrent Neural Network Grammars (RNNG)~\cite{dyer-EtAl:2016:N16-1}\footnote{The output of RNNG is not in descending order because it samples candidates based on their scores. Thus, we prepared more candidates (i.e., 700) to be able to obtain correct instances as candidates.}, that achieved 91.2 F1 score in our configuration.
Moreover, the lowest row of the middle part indicates the result by reranking the candidates from the retrained neural encoder-decoder based parser~\cite{P18-2097}.
Our base parser has two different parts from \newcite{P18-2097}.
First, we used the sum of the hidden states of the forward and backward RNNs as the hidden layer for each RNN\footnote{We used the deep bidirectional encoder described at \href{http://opennmt.net/OpenNMT/training/models/}{http://opennmt.net/OpenNMT/training/models/} instead of a basic bidirectional encoder.}.
Second, we tied the embedding matrix to the weight matrix to compute the probability distributions in the decoder.
The retrained parser achieved 93.12 F1 score.
Finally, we achieved 94.47 F1 score by reranking its candidates with AWD-LSTM-DOC.
We expect that we can achieve even better score by replacing the base parser with the current state-of-the-art one~\cite{P18-1249}.

\section{Related Work}
\newcite{Bengio:2003:NPL:944919.944966} are pioneers of neural language models.
To address the curse of dimensionality in language modeling, they proposed a method using word embeddings and a feed-forward neural network (FFNN).
They demonstrated that their approach outperformed n-gram language models, but FFNN can only handle fixed-length contexts.
Instead of FFNN, \newcite{DBLP:conf/interspeech/MikolovKBCK10} applied RNN~\cite{elman1990finding} to language modeling to address the entire given sequence as a context.
Their method outperformed the Kneser-Ney smoothed 5-gram language model~\cite{Kneser1995,Chen:1996:ESS:981863.981904}.

Researchers continue to try to improve the performance of RNN language models.
\newcite{DBLP:journals/corr/ZarembaSV14} used LSTM~\cite{Hochreiter:1997:LSM:1246443.1246450} instead of a simple RNN for language modeling and significantly improved an RNN language model by applying dropout~\cite{Srivastava:2014:DSW:2627435.2670313} to all the connections except for the recurrent connections.
To regularize the recurrent connections, \newcite{Gal2016Theoretically} proposed variational inference-based dropout.
Their method uses the same dropout mask at each timestep.
\newcite{fraternal} proposed fraternal dropout, which minimizes the differences between outputs from different dropout masks to be invariant to the dropout mask.
\newcite{DBLP:journals/corr/MelisDB17} used black-box optimization to find appropriate hyperparameters for RNN language models and demonstrated that the standard LSTM with proper regularizations can outperform other architectures.

Apart from dropout techniques, \newcite{DBLP:journals/corr/InanKS16} and \newcite{press-wolf:2017:EACLshort} proposed the word tying method (WT), which unifies word embeddings ($E$ in Equation \ref{eq:embed}) with the weight matrix to compute probability distributions ($W$ in Equation \ref{eq:softmax}).
In addition to quantitative evaluation, \newcite{DBLP:journals/corr/InanKS16} provided a theoretical justification for WT and proposed the augmented loss technique (AL), which computes an objective probability based on word embeddings.
In addition to these regularization techniques, \newcite{merityRegOpt} used DropConnect~\cite{wan2013regularization} and averaged SGD~\cite{polyak1992acceleration} for an LSTM language model.
Their AWD-LSTM achieved lower perplexity than \newcite{DBLP:journals/corr/MelisDB17} on PTB and WikiText-2.

Previous studies also explored superior architecture for language modeling.
\newcite{zilly2016recurrent} proposed recurrent highway networks that use highway layers~\cite{DBLP:journals/corr/SrivastavaGS15} to deepen recurrent connections.
\newcite{45826} adopted reinforcement learning to construct the best RNN structure.
However, as mentioned, \newcite{DBLP:journals/corr/MelisDB17} established that the standard LSTM is superior to these architectures.
Apart from RNN architecture, \newcite{takase-suzuki-nagata:2017:I17-2} proposed the input-to-output gate (IOG), which boosts the performance of trained language models.

As described in Section \ref{sec:mos}, \newcite{DBLP:journals/corr/abs-1711-03953} interpreted training language modeling as matrix factorization and improved performance by computing multiple probability distributions.
In this study, we generalized their approach to use the middle layers of RNNs.
Finally, our proposed method, DOC, achieved the state-of-the-art score on the standard benchmark datasets.

Some studies provided methods that boost performance by using statistics obtained from test data.
\newcite{DBLP:journals/corr/GraveJU16} extended a cache model~\cite{cache4ngram} for RNN language models.
\newcite{DBLP:journals/corr/abs-1709-07432} proposed dynamic evaluation that updates parameters based on a recent sequence during testing.
Although these methods might also improve the performance of DOC, we omitted such investigation to focus on comparisons among methods trained only on the training set.

\section{Conclusion}
We proposed \textit{Direct Output Connection} (DOC), a generalization method of MoS introduced by \newcite{DBLP:journals/corr/abs-1711-03953}.
DOC raises the expressive power of RNN language models and improves quality of the model.
DOC outperformed MoS and achieved the best perplexities on the standard benchmark datasets of language modeling: PTB and WikiText-2.
Moreover, we investigated its effectiveness on machine translation and headline generation.
Our results show that DOC also improved the performance of EncDec and using a middle layer positively affected such application tasks.

\bibliography{emnlp2018.bbl}

\begin{thebibliography}{45}
\expandafter\ifx\csname natexlab\endcsname\relax\def\natexlab#1{#1}\fi

\bibitem[{Bengio et~al.(2003)Bengio, Ducharme, Vincent, and
  Janvin}]{Bengio:2003:NPL:944919.944966}
Yoshua Bengio, R{\'e}jean Ducharme, Pascal Vincent, and Christian Janvin. 2003.
\newblock A neural probabilistic language model.
\newblock \emph{Journal of Machine Learning Research}, 3:1137--1155.

\bibitem[{Charniak(2000)}]{A00-2018}
Eugene Charniak. 2000.
\newblock A maximum-entropy-inspired parser.
\newblock In \emph{1st Meeting of the North American Chapter of the Association
  for Computational Linguistics (NAACL 2000)}, pages 132--139.

\bibitem[{Chen and Goodman(1996)}]{Chen:1996:ESS:981863.981904}
Stanley~F. Chen and Joshua Goodman. 1996.
\newblock An empirical study of smoothing techniques for language modeling.
\newblock In \emph{Proceedings of the 34th Annual Meeting on Association for
  Computational Linguistics (ACL 1996)}, pages 310--318.

\bibitem[{Choe and Charniak(2016)}]{choe-charniak:2016:EMNLP2016}
Do~Kook Choe and Eugene Charniak. 2016.
\newblock Parsing as language modeling.
\newblock In \emph{Proceedings of the 2016 Conference on Empirical Methods in
  Natural Language Processing (EMNLP 2016)}, pages 2331--2336.

\bibitem[{Dyer et~al.(2016)Dyer, Kuncoro, Ballesteros, and
  Smith}]{dyer-EtAl:2016:N16-1}
Chris Dyer, Adhiguna Kuncoro, Miguel Ballesteros, and Noah~A. Smith. 2016.
\newblock Recurrent neural network grammars.
\newblock In \emph{Proceedings of the 2016 Conference of the North American
  Chapter of the Association for Computational Linguistics: Human Language
  Technologies (NAACL-HLT 2016)}, pages 199--209.

\bibitem[{Elman(1990)}]{elman1990finding}
Jeffrey~L Elman. 1990.
\newblock {Finding Structure in Time}.
\newblock \emph{Cognitive science}, 14(2):179--211.

\bibitem[{Fried et~al.(2017)Fried, Stern, and
  Klein}]{fried-stern-klein:2017:Short}
Daniel Fried, Mitchell Stern, and Dan Klein. 2017.
\newblock Improving neural parsing by disentangling model combination and
  reranking effects.
\newblock In \emph{Proceedings of the 55th Annual Meeting of the Association
  for Computational Linguistics (ACL 2017)}, pages 161--166.

\bibitem[{Gal and Ghahramani(2016)}]{Gal2016Theoretically}
Yarin Gal and Zoubin Ghahramani. 2016.
\newblock {A Theoretically Grounded Application of Dropout in Recurrent Neural
  Networks}.
\newblock In \emph{Advances in Neural Information Processing Systems 29 (NIPS
  2016)}.

\bibitem[{Grave et~al.(2017)Grave, Joulin, and
  Usunier}]{DBLP:journals/corr/GraveJU16}
Edouard Grave, Armand Joulin, and Nicolas Usunier. 2017.
\newblock {Improving Neural Language Models with a Continuous Cache}.
\newblock In \emph{Proceedings of the 5th International Conference on Learning
  Representations (ICLR 2017)}.

\bibitem[{Hochreiter and
  Schmidhuber(1997)}]{Hochreiter:1997:LSM:1246443.1246450}
Sepp Hochreiter and J\"{u}rgen Schmidhuber. 1997.
\newblock {Long Short-Term Memory}.
\newblock \emph{Neural Computation}, 9(8):1735--1780.

\bibitem[{Inan et~al.(2017)Inan, Khosravi, and
  Socher}]{DBLP:journals/corr/InanKS16}
Hakan Inan, Khashayar Khosravi, and Richard Socher. 2017.
\newblock {Tying Word Vectors and Word Classifiers: {A} Loss Framework for
  Language Modeling}.
\newblock In \emph{Proceedings of the 5th International Conference on Learning
  Representations (ICLR 2017)}.

\bibitem[{Kitaev and Klein(2018)}]{P18-1249}
Nikita Kitaev and Dan Klein. 2018.
\newblock Constituency parsing with a self-attentive encoder.
\newblock In \emph{Proceedings of the 56th Annual Meeting of the Association
  for Computational Linguistics (ACL 2018)}, pages 2676--2686.

\bibitem[{Kiyono et~al.(2017)Kiyono, Takase, Suzuki, Okazaki, Inui, and
  Nagata}]{kiyono}
Shun Kiyono, Sho Takase, Jun Suzuki, Naoaki Okazaki, Kentaro Inui, and Masaaki
  Nagata. 2017.
\newblock Source-side prediction for neural headline generation.
\newblock \emph{CoRR}.

\bibitem[{Kneser and Ney(1995)}]{Kneser1995}
Reinhard Kneser and Hermann Ney. 1995.
\newblock Improved backing-off for m-gram language modeling.
\newblock In \emph{In Proceedings of the IEEE International Conference on
  Acoustics, Speech and Signal Processing (ICASSP 1995)}, pages 181--184.

\bibitem[{Krause et~al.(2017)Krause, Kahembwe, Murray, and
  Renals}]{DBLP:journals/corr/abs-1709-07432}
Ben Krause, Emmanuel Kahembwe, Iain Murray, and Steve Renals. 2017.
\newblock Dynamic evaluation of neural sequence models.
\newblock \emph{CoRR}.

\bibitem[{Kuhn and De~Mori(1990)}]{cache4ngram}
Roland Kuhn and Renato De~Mori. 1990.
\newblock {A} cache-based natural language model for speech recognition.
\newblock 12:570--583.

\bibitem[{Luong et~al.(2015)Luong, Pham, and
  Manning}]{luong-pham-manning:2015:EMNLP}
Thang Luong, Hieu Pham, and Christopher~D. Manning. 2015.
\newblock Effective approaches to attention-based neural machine translation.
\newblock In \emph{Proceedings of the 2015 Conference on Empirical Methods in
  Natural Language Processing (EMNLP 2015)}, pages 1412--1421.

\bibitem[{Marcus et~al.(1993)Marcus, Marcinkiewicz, and
  Santorini}]{Marcus:1993:BLA:972470.972475}
Mitchell~P. Marcus, Mary~Ann Marcinkiewicz, and Beatrice Santorini. 1993.
\newblock {Building a Large Annotated Corpus of English: The Penn Treebank}.
\newblock \emph{Computational Linguistics}, 19(2):313--330.

\bibitem[{Melis et~al.(2018)Melis, Dyer, and
  Blunsom}]{DBLP:journals/corr/MelisDB17}
G{\'{a}}bor Melis, Chris Dyer, and Phil Blunsom. 2018.
\newblock On the state of the art of evaluation in neural language models.
\newblock \emph{Proceedings of the 6th International Conference on Learning
  Representations (ICLR 2018)}.

\bibitem[{Merity et~al.(2018)Merity, Keskar, and Socher}]{merityRegOpt}
Stephen Merity, Nitish~Shirish Keskar, and Richard Socher. 2018.
\newblock {Regularizing and Optimizing LSTM Language Models}.
\newblock In \emph{Proceedings of the 6th International Conference on Learning
  Representations (ICLR 2018)}.

\bibitem[{Merity et~al.(2017)Merity, Xiong, Bradbury, and
  Socher}]{DBLP:journals/corr/MerityXBS16}
Stephen Merity, Caiming Xiong, James Bradbury, and Richard Socher. 2017.
\newblock {Pointer Sentinel Mixture Models}.
\newblock In \emph{Proceedings of the 5th International Conference on Learning
  Representations (ICLR 2017)}.

\bibitem[{Mikolov et~al.(2010)Mikolov, Karafi{\'{a}}t, Burget, Cernock{\'{y}},
  and Khudanpur}]{DBLP:conf/interspeech/MikolovKBCK10}
Tomas Mikolov, Martin Karafi{\'{a}}t, Luk{\'{a}}s Burget, Jan Cernock{\'{y}},
  and Sanjeev Khudanpur. 2010.
\newblock Recurrent neural network based language model.
\newblock In \emph{Proceedings of the 11th Annual Conference of the
  International Speech Communication Association (INTERSPEECH 2010)}, pages
  1045--1048.

\bibitem[{Mikolov et~al.(2013)Mikolov, Sutskever, Chen, Corrado, and
  Dean}]{NIPS2013_5021}
Tomas Mikolov, Ilya Sutskever, Kai Chen, Greg~S Corrado, and Jeff Dean. 2013.
\newblock {Distributed Representations of Words and Phrases and their
  Compositionality}.
\newblock In \emph{Advances in Neural Information Processing Systems 26 (NIPS
  2013)}, pages 3111--3119.

\bibitem[{Mnih and Kavukcuoglu(2013)}]{NIPS2013_5165}
Andriy Mnih and Koray Kavukcuoglu. 2013.
\newblock {Learning Word Embeddings Efficiently with Noise-Contrastive
  Estimation}.
\newblock In \emph{Advances in Neural Information Processing Systems 26 (NIPS
  2013)}, pages 2265--2273.

\bibitem[{Napoles et~al.(2012)Napoles, Gormley, and
  Van~Durme}]{Napoles:2012:AG:2391200.2391218}
Courtney Napoles, Matthew Gormley, and Benjamin Van~Durme. 2012.
\newblock Annotated gigaword.
\newblock In \emph{Proceedings of the Joint Workshop on Automatic Knowledge
  Base Construction and Web-scale Knowledge Extraction}, pages 95--100.

\bibitem[{Peters et~al.(2018)Peters, Neumann, Iyyer, Gardner, Clark, Lee, and
  Zettlemoyer}]{N18-1202}
Matthew Peters, Mark Neumann, Mohit Iyyer, Matt Gardner, Christopher Clark,
  Kenton Lee, and Luke Zettlemoyer. 2018.
\newblock Deep contextualized word representations.
\newblock In \emph{Proceedings of the 2018 Conference of the North American
  Chapter of the Association for Computational Linguistics: Human Language
  Technologies (NAACL 2018)}, pages 2227--2237.

\bibitem[{Polyak and Juditsky(1992)}]{polyak1992acceleration}
Boris~T Polyak and Anatoli~B Juditsky. 1992.
\newblock {Acceleration of Stochastic Approximation by Averaging}.
\newblock \emph{SIAM Journal on Control and Optimization}, 30(4):838--855.

\bibitem[{Press and Wolf(2017)}]{press-wolf:2017:EACLshort}
Ofir Press and Lior Wolf. 2017.
\newblock {Using the Output Embedding to Improve Language Models}.
\newblock In \emph{Proceedings of the 15th Conference of the European Chapter
  of the Association for Computational Linguistics (EACL 2017)}, pages
  157--163.

\bibitem[{Rush et~al.(2015)Rush, Chopra, and
  Weston}]{rush-chopra-weston:2015:EMNLP}
Alexander~M. Rush, Sumit Chopra, and Jason Weston. 2015.
\newblock {A Neural Attention Model for Abstractive Sentence Summarization}.
\newblock In \emph{Proceedings of the 2015 Conference on Empirical Methods in
  Natural Language Processing (EMNLP 2015)}, pages 379--389.

\bibitem[{Sennrich et~al.(2016)Sennrich, Haddow, and
  Birch}]{sennrich-haddow-birch:2016:P16-11}
Rico Sennrich, Barry Haddow, and Alexandra Birch. 2016.
\newblock Improving neural machine translation models with monolingual data.
\newblock In \emph{Proceedings of the 54th Annual Meeting of the Association
  for Computational Linguistics (ACL 2016)}, pages 86--96.

\bibitem[{Shazeer et~al.(2017)Shazeer, Mirhoseini, Maziarz, Davis, Le, Hinton,
  and Dean}]{DBLP:journals/corr/ShazeerMMDLHD17}
Noam Shazeer, Azalia Mirhoseini, Krzysztof Maziarz, Andy Davis, Quoc~V. Le,
  Geoffrey~E. Hinton, and Jeff Dean. 2017.
\newblock Outrageously large neural networks: The sparsely-gated
  mixture-of-experts layer.
\newblock In \emph{Proceedings of the 5th International Conference on Learning
  Representations (ICLR 2017)}.

\bibitem[{Srivastava et~al.(2014)Srivastava, Hinton, Krizhevsky, Sutskever, and
  Salakhutdinov}]{Srivastava:2014:DSW:2627435.2670313}
Nitish Srivastava, Geoffrey Hinton, Alex Krizhevsky, Ilya Sutskever, and Ruslan
  Salakhutdinov. 2014.
\newblock Dropout: A simple way to prevent neural networks from overfitting.
\newblock \emph{Journal of Machine Learning Research}, 15(1):1929--1958.

\bibitem[{Srivastava et~al.(2015)Srivastava, Greff, and
  Schmidhuber}]{DBLP:journals/corr/SrivastavaGS15}
Rupesh~Kumar Srivastava, Klaus Greff, and J{\"{u}}rgen Schmidhuber. 2015.
\newblock Highway networks.
\newblock In \emph{Proceedings of the Deep Learning Workshop in ICML 15}.

\bibitem[{Sutskever et~al.(2014)Sutskever, Vinyals, and
  Le}]{Sutskever:2014:SSL:2969033.2969173}
Ilya Sutskever, Oriol Vinyals, and Quoc~V. Le. 2014.
\newblock {Sequence to Sequence Learning with Neural Networks}.
\newblock In \emph{Advances in Neural Information Processing Systems 27 (NIPS
  2014)}, pages 3104--3112.

\bibitem[{Suzuki et~al.(2018)Suzuki, Takase, Kamigaito, Morishita, and
  Nagata}]{P18-2097}
Jun Suzuki, Sho Takase, Hidetaka Kamigaito, Makoto Morishita, and Masaaki
  Nagata. 2018.
\newblock An empirical study of building a strong baseline for constituency
  parsing.
\newblock In \emph{Proceedings of the 56th Annual Meeting of the Association
  for Computational Linguistics (ACL 2018)}, pages 612--618.

\bibitem[{Szegedy et~al.(2015)Szegedy, Liu, Jia, Sermanet, Reed, Anguelov,
  Erhan, Vanhoucke, and Rabinovich}]{43022}
Christian Szegedy, Wei Liu, Yangqing Jia, Pierre Sermanet, Scott Reed, Dragomir
  Anguelov, Dumitru Erhan, Vincent Vanhoucke, and Andrew Rabinovich. 2015.
\newblock Going deeper with convolutions.
\newblock In \emph{Proceedings of the 28th IEEE Conference on Computer Vision
  and Pattern Recognition (CVPR 2015)}, pages 1--9.

\bibitem[{Takase et~al.(2017)Takase, Suzuki, and
  Nagata}]{takase-suzuki-nagata:2017:I17-2}
Sho Takase, Jun Suzuki, and Masaaki Nagata. 2017.
\newblock Input-to-output gate to improve rnn language models.
\newblock In \emph{Proceedings of the Eighth International Joint Conference on
  Natural Language Processing (IJCNLP 2017)}, pages 43--48.

\bibitem[{Wan et~al.(2013)Wan, Zeiler, Zhang, Cun, and
  Fergus}]{wan2013regularization}
Li~Wan, Matthew Zeiler, Sixin Zhang, Yann~L Cun, and Rob Fergus. 2013.
\newblock {Regularization of Neural Networks using DropConnect}.
\newblock In \emph{Proceedings of the 30th International Conference on Machine
  Learning (ICML 2013)}, pages 1058--1066.

\bibitem[{Wen et~al.(2015)Wen, Gasic, Mrk\v{s}i\'{c}, Su, Vandyke, and
  Young}]{wen-EtAl:2015:EMNLP}
Tsung-Hsien Wen, Milica Gasic, Nikola Mrk\v{s}i\'{c}, Pei-Hao Su, David
  Vandyke, and Steve Young. 2015.
\newblock {Semantically Conditioned LSTM-based Natural Language Generation for
  Spoken Dialogue Systems}.
\newblock In \emph{Proceedings of the 2015 Conference on Empirical Methods in
  Natural Language Processing (EMNLP 2015)}, pages 1711--1721.

\bibitem[{Yang et~al.(2018)Yang, Dai, Salakhutdinov, and
  Cohen}]{DBLP:journals/corr/abs-1711-03953}
Zhilin Yang, Zihang Dai, Ruslan Salakhutdinov, and William~W. Cohen. 2018.
\newblock Breaking the softmax bottleneck: {A} high-rank {RNN} language model.
\newblock In \emph{Proceedings of the 6th International Conference on Learning
  Representations (ICLR 2018)}.

\bibitem[{Zaremba et~al.(2014)Zaremba, Sutskever, and
  Vinyals}]{DBLP:journals/corr/ZarembaSV14}
Wojciech Zaremba, Ilya Sutskever, and Oriol Vinyals. 2014.
\newblock Recurrent neural network regularization.
\newblock In \emph{Proceedings of the 2nd International Conference on Learning
  Representations (ICLR 2014)}.

\bibitem[{Zhou et~al.(2017)Zhou, Yang, Wei, and Zhou}]{zhou-EtAl:2017:Long}
Qingyu Zhou, Nan Yang, Furu Wei, and Ming Zhou. 2017.
\newblock Selective encoding for abstractive sentence summarization.
\newblock In \emph{Proceedings of the 55th Annual Meeting of the Association
  for Computational Linguistics (ACL 2017)}, pages 1095--1104.

\bibitem[{Zilly et~al.(2017)Zilly, Srivastava, Koutn{\'\i}k, and
  Schmidhuber}]{zilly2016recurrent}
Julian~Georg Zilly, Rupesh~Kumar Srivastava, Jan Koutn{\'\i}k, and J{\"u}rgen
  Schmidhuber. 2017.
\newblock {Recurrent Highway Networks}.
\newblock \emph{Proceedings of the 34th International Conference on Machine
  Learning (ICML 2017)}, pages 4189--4198.

\bibitem[{Zolna et~al.(2018)Zolna, Arpit, Suhubdy, and Bengio}]{fraternal}
Konrad Zolna, Devansh Arpit, Dendi Suhubdy, and Yoshua Bengio. 2018.
\newblock Fraternal dropout.
\newblock In \emph{Proceedings of the 6th International Conference on Learning
  Representations (ICLR 2018)}.

\bibitem[{Zoph and Le(2017)}]{45826}
Barret Zoph and Quoc~V. Le. 2017.
\newblock {Neural Architecture Search with Reinforcement Learning}.
\newblock In \emph{Proceedings of the 5th International Conference on Learning
  Representations (ICLR 2017)}.

\end{thebibliography}
\bibliographystyle{acl_natbib_nourl}

\end{document}